\documentclass[10pt,twocolumn,letterpaper]{article}
\usepackage{cvpr}              

\usepackage{graphicx}
\usepackage{amsmath}
\usepackage{amssymb}
\usepackage{booktabs}
\usepackage{multirow}
\usepackage{multicol}
\usepackage{bbding}
\usepackage{xr}
\usepackage[pagebackref,breaklinks,colorlinks]{hyperref}
\usepackage{appendix}
\usepackage[capitalize]{cleveref}
\crefname{section}{Sec.}{Secs.}
\Crefname{section}{Section}{Sections}
\Crefname{table}{Table}{Tables}
\crefname{table}{Tab.}{Tabs.}

\def\cvprPaperID{17102} 
\def\confName{CVPR}
\def\confYear{2024}

\begin{document}

\title{Natural-language-driven Simulation Benchmark and Copilot for\\Efficient Production of Object Interactions in Virtual Road Scenes}

\author{Kairui Yang\thanks{Co-first authors.}, Zihao Guo\footnotemark[1], Gengjie Lin, Haotian Dong, Die Zuo, Jibin Peng, \\Zhao Huang, Zhecheng Xu, Fupeng Li, Ziyun Bai, Di Lin\thanks{Corresponding authors.}\\
\\ 
Tianjin University\\
}

\maketitle

\begin{abstract}


We advocate the idea of the natural-language-driven (NLD) simulation to efficiently produce the object interactions between multiple objects in the virtual road scenes, for teaching and testing the autonomous driving systems that should take quick action to avoid collision with obstacles with unpredictable motions. The NLD simulation allows the brief natural-language description to control the object interactions, significantly reducing the human efforts for creating a large amount of interaction data. To facilitate the research of NLD simulation, we collect the Language-to-Interaction (L2I) benchmark dataset with 120,000 natural-language descriptions of object interactions in 6 common types of road topologies. Each description is associated with the programming code, which the graphic render can use to visually reconstruct the object interactions in the virtual scenes. As a methodology contribution, we design SimCopilot to translate the interaction descriptions to the renderable code. We use the L2I dataset to evaluate SimCopilot's abilities to control the object motions, generate complex interactions, and generalize interactions across road topologies. The L2I dataset and the evaluation results motivate the relevant research of the NLD simulation.

\end{abstract} 
\section{Introduction}

The rapid growth of modern cities gave birth to the largest-scale road system to date in the history of humankind. Fast-moving cars and non-vehicle objects (e.g., pedestrians and bicycles) exist in the contemporary road system. Wisely understanding the interaction between the objects in the road scenes, which consists of the motion sequences of multiple objects, is essential to constructing the intelligent road system, which enables applications like self-driving vehicles and intelligent control of traffic flow. Note that object interactions vary according to the object instances and categories co-existing in different roads, making understanding the object interactions challenging.

People get used to collecting road data from real-world scenes for training deep networks, which shows a solid ability to understand object interactions. Typically, this is done by employing in-car or street-view cameras to capture the objects of interest in different scenarios, whose interactions are tracked and recorded over time. However, most of the scenarios captured by cameras are average. It means the long-tailed interactions in some critical scenarios relevant to the object collision are complex. Intuitively, without sufficient long-tailed interactions, we lack the adversarial samples for training the deep networks, thus failing to facilitate a more robust system to lower the risk of traffic accidents.

\begin{figure}[t!]
\centering
\includegraphics[width=\linewidth]{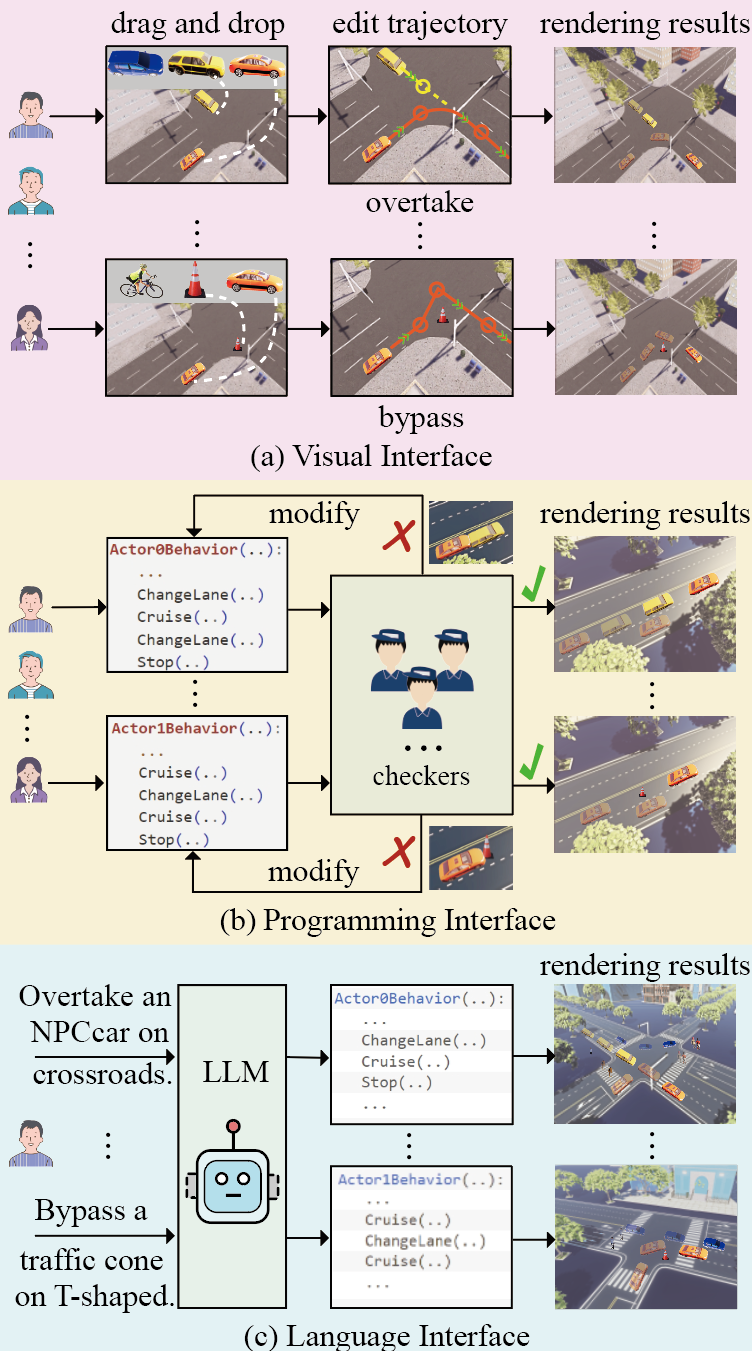}
\vspace{-0.25in}
\caption{The comparison between the visual, programming, and language interfaces for simulating the object interactions in the virtual road scenes.}
\vspace{-0.3in}
\label{fig:teaser}
\end{figure}

The popular manner employs the simulation engine to create long-tailed interactions between objects in the virtual roads. As illustrated in Figure~\ref{fig:teaser}(a--b), the broadly used simulation engines allow the user to control the objects via the visual~\cite{apollo, nvidia, shah2018airsim, samak2023autodrive, dosovitskiy2017carla, rong2020lgsvl, silvera2022dreye, rong2020lgsvl, dosovitskiy2017carla} or programming interfaces~\cite{rempe2023trace, janner2022planning, zhong2023guided, fremont2022scenic, openscenario, li2019adaps, shah2018airsim, dosovitskiy2017carla, zhang2021end, rong2020lgsvl, amini2020learning}, where the latter enables higher flexibility for managing the motion properties of the object motions (e.g., waypoint, heading, and velocity). However, using the programming language for simulation also involves a high complexity in the simulation of object interactions, especially where various objects emerge with complicated motions.

This paper promotes a relatively brand-new idea of the natural-language-driven (NLD) simulation for producing object interactions in virtual roads. This can be done by building a language interface, which uses the concise description of object interactions in the natural language to manage the simulation (see Figure~\ref{fig:teaser}(c)). In contrast to the visual and programable simulation, the language interface can take advantage of the recent progress on the large language models (LLMs)~\cite{kenton2019bert, stiennon2020learning, brown2020language, touvron2023llama, qwen} from two aspects. First, the natural language provides high-level semantic information for significantly simplifying the simulation process. It can depict complex scenarios where multiple objects compete for the road space dynamically, rather than needing many manual operations or code lines to delineate each object's motion sequence. Second, the text description of object interactions can be easily generalized across variant road topologies by slightly modifying the text prompt for the LLM. In comparison, the visual and programmable simulations need to carefully tune the object motion specific to a road topology to avoid unreasonable motion (e.g., the car driving out of the road boundary) or missing critical interaction (e.g., the far-way cars without competition for the road space) across various road topologies.

We advocate a kind of LLM as the simulation copilot (SimCopiloit) to complete the NLD simulation. SimCopilot translates the natural-language descriptions of the object interactions, outputting the programming-language codes for the simulation engine. Because natural-language descriptions can be achieved efficiently, while the programming codes can be used to control the details of object motions in the simulated scenes, SimCopilot can take advantage of both worlds of efficiency and effectiveness. We contribute the \emph{\textbf{Language-to-Interaction}} (L2I) benchmark dataset to the research community for investigating SimCopilot and its variants. As illustrated in Figure~\ref{fig:dataset}, L2I provides six road topologies (i.e., straightway, bend, roundabout, cross intersection, T-shaped intersection, and Y-shaped intersection). In each road topology, we allow SimCopilot to control an ego car to interact with typical obstacles (i.e., NPC cars, pedestrians, bicycles, and traffic cones). We prepare 120,000 scenarios in the L2I dataset.

L2I offers challenging data for researchers to train and evaluate variants of SimCopilot in the following three tasks to enable real-world applications. First, L2I describes the motion parameters of the ego car and obstacles in the natural language. We pair the natural-language description with the code segment for simulation. One can use the paired data to train and test SimCopilot's capacity to translate the motions from natural to programming language. Second, the ego car may co-exist with the static and moving obstacles in various kinds and numbers. A brilliant language model, well-trained on the simple interaction between the ego car and fewer obstacles, can be extended to generate the intricate interaction between more objects. Third, the same types of road topologies in the training and testing sets of L2I show various geometries, such as different numbers and shapes of lanes. By training and testing the language model of SimCopilot, we provide a chance to augment the generalization power of SimCopilot that produces object interactions across discrepant road topologies.

We extensively train and test the latest LLMs on the L2I dataset, analyzing the performances of these LLMs adapted to the task of the NLD simulation. Our analysis can be regarded as the baseline for the comparison in future works. Below, we brief our contributions as:
\begin{itemize}
\item We advocate a novel idea of using concise natural language to simulate the object interactions in the road scenes. The NLD simulation has a great potency to significantly reduce the effort of constructing road scenes where complex object interactions emerge.
\item We collect the L2I dataset to benchmark the NLD simulation. L2I benchmarks the tasks of translating the object motions from the natural to programming language, producing complex object interactions, and generalizing them across different road topologies. It hopefully advances the real-world applications of the NLD simulation.
\item We adapt the recent LLMs to the NLD simulation, forming a baseline SimCopilot, whose performances are comprehensively evaluated on the L2I dataset. The experimental results can motivate future investigation on the NLD simulation.
\end{itemize}


\section{Related Work}



Below, we introduce three kinds of simulators according to interfaces of controlling the object interactions.

\vspace{0.05in}
\noindent{\bf Visual Interface~~}
The simulators with visual interfaces can be divided into Logsim and Worldsim. Logsim~\cite{apollo} uses the real scenes to provide complex traffic situations. The scenes provided by Logsim are unchangeable, thus it fails to test the object interactions specified by different users. Worldsim allows users to design the scenes for testing the autonomous vehicles. These simulators~\cite{apollo, nvidia, shah2018airsim, samak2023autodrive, dosovitskiy2017carla, rong2020lgsvl, silvera2022dreye} employ the game engines like Unreal~\cite{ue4} and Unity 3D~\cite{ue} that have the graphic renders of the realistic scenes obeying the real-world rules. Nevertheless, Logsim costs expensive labor to operate the visual interface to control every object to create object interactions in the road scene. In contrast, the language interface relies on the short text of natural language to efficiently create the object interactions.

\vspace{0.05in}
\noindent{\bf Programming Interface~~}
Compared to the visual interface that partially omits the details of object motions to simplify the simulation, the programming interface offers an extensive collection of controlling functions of the object interactions. Scenic~\cite{fremont2022scenic} and OpenSCENARIO~\cite{openscenario} are the domain-specific languages for describing the car behaviors. They are typical methods of using the formal language for simulation. Yet, they need long pages of code segments to build a road scene with complex object interactions. The reinforcement learning~\cite{rempe2023trace, janner2022planning, li2019adaps, shah2018airsim, dosovitskiy2017carla, zhang2021end, rong2020lgsvl, amini2020learning, zhong2023guided} allows the traffic rules, which also belong to the formal language, to guide the simulation of the specific object interactions without requiring a fussy coding work. The traffic rules help to construct the reward function that drives reinforcement learning. However, the limited set of traffic rules indirectly describes the details of object motions, lacking a solid control of the desired interactions. In contrast, various kinds of motion details can be easily added to the description in natural language for controlling the object interactions.

\vspace{0.05in}
\noindent{\bf Language Interface~~}
The recent LLMs~\cite{kenton2019bert, stiennon2020learning, brown2020language, touvron2023llama, qwen} enable the language interface to create virtual scenes with complicated object interactions. The text-to-image methods~\cite{liu2023instaflow, ramesh2021zero, ramesh2022hierarchical, NEURIPS2022_ec795aea, ho2022imagen} adjust the object appearances in the generated images according to the natural-language description. Nevertheless, these methods inadequately capture the relationship between dynamic objects. Though people can use the text-to-video methods~\cite{Blattmann_2023_CVPR, wu2023tune, ho2022video, voleti2022mcvd, runway} and traffic simulation methods~\cite{park2023generative, zhong2023language, zhong2023guided, OpenAI2023GPT4TR, vemprala2023chatgpt,cui2023drive, wen2023dilu, chen2023driving, mao2023gpt, sha2023languagempc, jin2023surrealdriver} to design the object interactions, these methods may fail to produce the usual object motions like those in the real world. CTG~\cite{zhong2023guided} and CTG++~\cite{zhong2023language} depend on the traffic rules to generate realistic motions for the individual objects, yet lack a practical scheme for producing the complex object interactions. It motivates a better means for creating more natural and complex object interactions.

\section{Language-to-Interaction Dataset}


We introduce the road scenes in the L2I dataset and how to produce them via the simulation engine.

\begin{figure*}[t!]
\centering
\includegraphics[width=\linewidth]{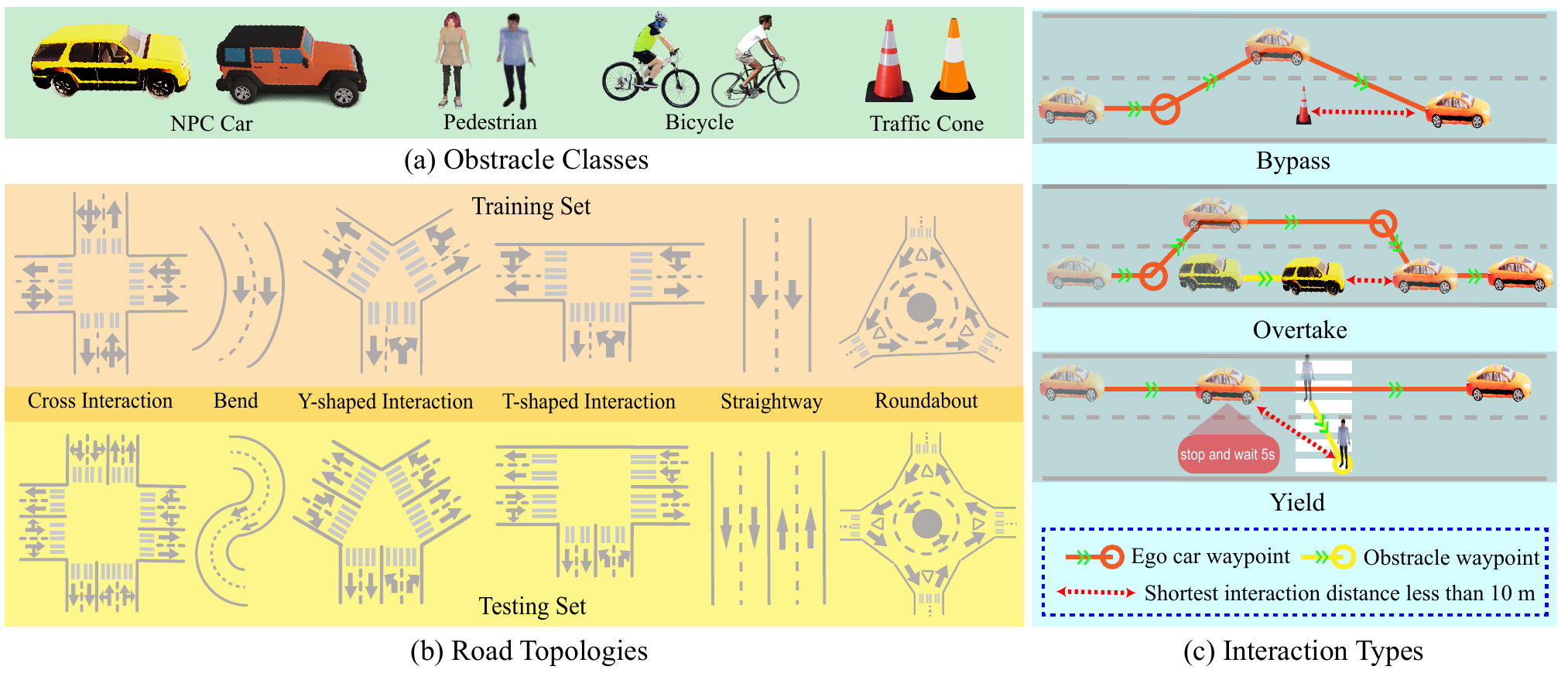}
\vspace{-0.3in}
\caption{The basic information of the L2I dataset. (a) Apart from the ego car, we prepare four kinds of obstacles including the NPC car, pedestrian, bicycle, and traffic cone. (b) There are six kinds of road topologies, where the road shapes and the numbers of lanes in the training and testing sets are different. (c) Bypassing the static obstacle, overtaking and yielding the dynamic obstacle are the three interactions possibly appearing in the L2I dataset.}
\label{fig:dataset}
\vspace{-0.15in}
\end{figure*}

\subsection{Basic Information of L2I}
\label{subsec:basic}

The L2I dataset contains 120,000 virtual road scenes, where the obstacle classes, road topologies, and interactions are illustrated in Figure~\ref{fig:dataset}. More examples of road scenes with short-range interactions can be found in Section 1 of the supplementary file.

\vspace{0.05in}
\noindent{\bf Obstacle Classes~~}
An ego car interacts with at least an obstacle in the road scene. Each obstacle can be taken from the classes of NPC car, pedestrian, bicycle, and traffic cone (see Figure~\ref{fig:dataset}(a)), which are widely seen in reality. Expect the traffic cones that are always static; other obstacles can be stationary or moving. The moving ego car or obstacle has the sequential motions of cruising, changing lanes, and stopping. Cruising or changing lanes means moving along the same lane or across different lanes. The motion of stopping can be temporary or permanent.

\vspace{0.05in}
\noindent{\bf Road Topologies~~}
As illustrated in Figure~\ref{fig:dataset}(b), we prepare six typical kinds of road topology in the L2I dataset, including straightway, bend, roundabout, cross intersection, T-shaped intersection, and Y-shaped intersection. Each kind of road topology has two variants with discrepant shapes and lanes. Each variant appears in 10,000 scenes. We use the XODR file to store each road topology with multiple lanes. A set of 2D coordinates represents the boundary of each lane. Following these lanes, the ego car can choose a goal of entering/exiting the topology, going straight, or turning left/right. We change the shapes and numbers of the lanes in the training and testing sets.

\vspace{0.05in}
\noindent{\bf Interaction Types~~}
In each scene, the ego car interacts with 1$\sim$5 static/moving obstacles. Each number of obstacles corresponds to 24,000 scenes. In contrast to the motion belonging to the single object, the interaction defined here captures the relationship between the ego car and an obstacle. The possible interactions present in the L2I dataset are illustrated in Figure~\ref{fig:dataset}(c). The ego car can bypass a static obstacle that occupies the road space. During bypassing, the ego car changes lanes to avoid collision with the stationary obstacle. The ego car can also overtake or yield the moving obstacle. During overtaking or yielding, the ego car occupies a lane sooner or later than the obstacle.

We investigate the high-intensity interactions between the ego car and obstacles. These high-intensity interactions should be completed in a short range of less than 10 meters. In contrast, the long-range interactions are less likely to do with most long-tailed scenarios, where the ego car must take prompt action to avoid collision with the obstacles. We omit too long-range interactions in the dataset, thus facilitating a more focused study of the critical object interactions.

\subsection{Production of Virtual Road Scenes}

We redevelop a simulation engine based on the SVL simulator~\cite{svlsimulator} to produce the data in the L2I dataset.

\vspace{0.05in}
\noindent{\bf Editing of Road Scenes~~}
The SVL simulator has a visual editor, which allows the user to manually edit the positions and motion trajectories of the ego car and obstacles by respecting their interactions in the same road scene. It should be noted that the editing is done from the bird's eye view.

\begin{figure*}[t!]
\centering
\includegraphics[width=\linewidth]{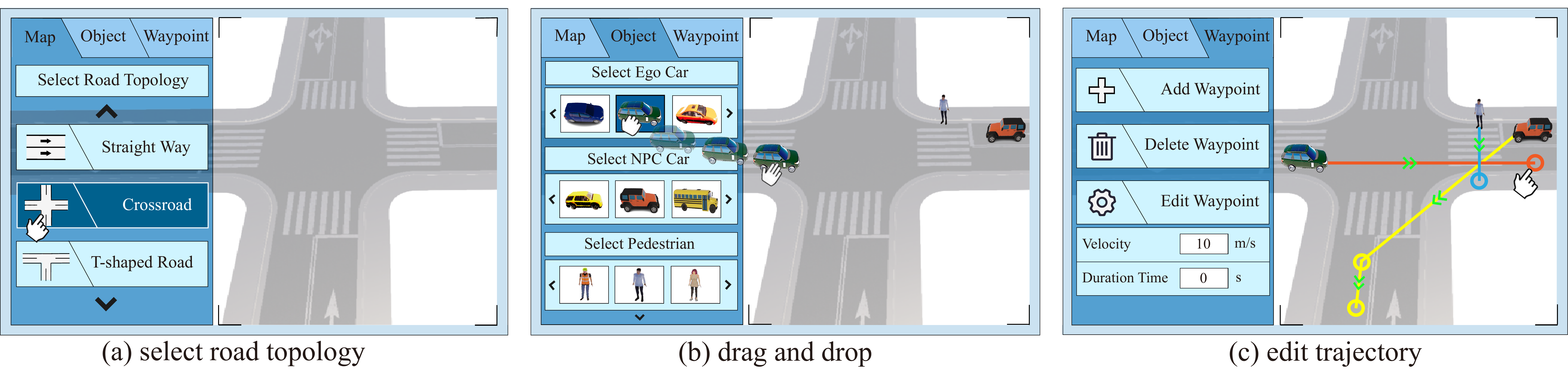}
\vspace{-0.25in}
\caption{Three steps for producing a virtual road scene. (a) First, we select the road topology from that can be regarded as a canvas. (b) Second, we drag and drop the ego car and obstacles of different categories onto the road topology. (c) Third, we edit the trajectory of every object on the road, where each waypoint (see the color circle) and the associated velocity/acceleration/duration/distance can be edited.}
\label{fig:production_scene}
\vspace{-0.2in}
\end{figure*}

The user edits a scene in three steps. First, the user selects one of the built-in road topologies from the pre-scribed menu (see Figure~\ref{fig:production_scene}(a)). Second, the ego car and obstacles are dragged and dropped onto the road (see Figure~\ref{fig:production_scene}(b)). This step determines the initial positions of these objects within the road's boundary. Each dropped object is assigned an index. Third, the user edits the trajectories of the ego car and obstacles. The trajectory of each object consists of a set of waypoints within the road (see Figure~\ref{fig:production_scene}(c)). The waypoints subdivide the trajectory into several segments. Each waypoint has the editable value of velocity. The first and last waypoints' velocities of the trajectory are enforced to 0. The velocities associated with the middle waypoints can be set to 0 for an editable duration, meaning the object can stop temporarily at the waypoint. The last waypoint of the trajectory has the maximum duration time permitted, letting the object stop permanently. Each segment between a pair of adjacent waypoints represents the direction and movement distance of the object. The difference between the velocities of the adjacent waypoints means the object uniformly accelerates/decelerates along the trajectory segment.

We employ 60 users who can fluently utilize the editor to build 120,000 virtual road scenes. Each user accounts for the interactions in 2,000 scenes. We require each user to produce the short-range interaction between the ego car and every obstacle in the same scene. This is done by editing the shortest distance between the waypoints of the ego car and each obstacle to be less than 10 meters without collision or conflicting with any
traffic rule. Within distances less than 10 meters, the user must let the ego car bypass/overtake/yield the obstacle within the road topology. We record the positions of the ego car and the obstacle, where the interaction is completed, as the ground truth for evaluation. The editor eventually outputs the JSON script to record the data of the ego car, obstacles, trajectory segments, and waypoints edited by the user.

\begin{figure}[t!]
\centering
\includegraphics[width=\linewidth]{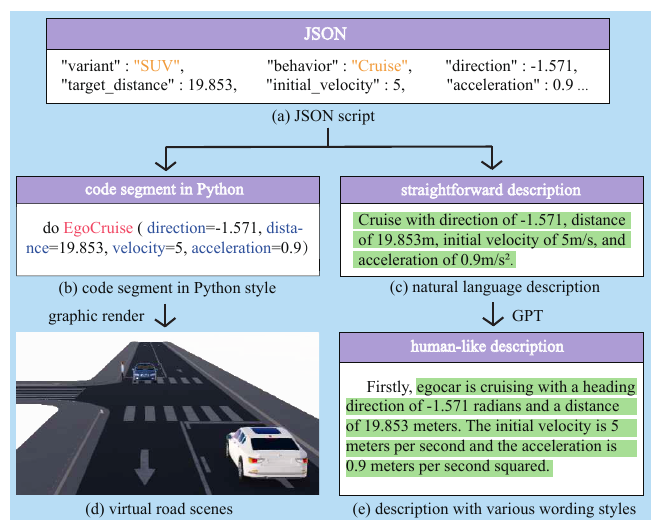}
\vspace{-0.2in}
\caption{The process of translating the JSON script of the virtual road scene to the code segment (see the left branch) and the natural-language description (see right branch).}
\label{fig:translate}
\vspace{-0.2in}
\end{figure}

\vspace{0.05in}
\noindent{\bf Code Segment for Simulation~~}
We build a programming-language interpreter to transform the JSON script to the code segment in Python (see the left branch of Figure~\ref{fig:translate}). The graphic render of the SVL simulator interprets the code segment and produces the road scene from the third-person, in-car, and unrestricted views. The code segment is mainly implemented by the functions of \emph{Cruise}, \emph{ChangeLane}, and \emph{Stop}. These functions are redeveloped on top of the Scenic-2 interface~\cite{fremont2022scenic} to control the object motions.

The programming-language interpreter transforms the trajectory segment within the identical lane or across different lanes into the function \emph{Cruise} or \emph{ChangeLane}. It should be noted that \emph{Cruise} and \emph{ChangeLane} have the parameters of \emph{direction}, \emph{distance}, \emph{initial velocity}, and \emph{accelerated velocity}. We compute \emph{direction} and \emph{distance} by respecting the direction and distance of the trajectory segment. The beginning waypoint of the trajectory segment contributes its velocity that plays as the parameter \emph{initial velocity} of the function, while the parameter \emph{accelerated velocity} is computed based on the trajectory segment's distance and the adjacent waypoints' velocities. The zero velocity of a waypoint is interpreted to the function \emph{Stop}, which has the parameter \emph{duration time} equal to the duration for the waypoint.

\vspace{0.05in}
\noindent{\bf Natural-language Description~~}
We develop an interpreter to transform the JSON script to the natural-language description of the object interactions (see the right branch of Figure~\ref{fig:translate}). This interpreter not only respects the trajectory segments and waypoints in the JSON script to organize the object motions of cruising, changing lanes, and stopping, thus forming a detailed description of the above motions. It also provides a summary of the object interactions.

Precisely, the detailed description of object motions consists of several paragraphs, where the sentences have the motion parameters of the objects. Figure~\ref{fig:production_scene}(c) shows an example of the description, where a sentence can be regarded as a relatively straightforward translation of the motion function into the natural language. However, this description is monotonous, losing the necessary diversity of human-like wording for training SimCopilot. To alleviate this problem, we employ the powerful dialogue-based language models, ChatGPT~\cite{stiennon2020learning, brown2020language} and Llama~\cite{touvron2023llama}, to rewrite the above description with various wording styles.

Based on the JSON script, the interpreter outputs the summary of the road scene for recording the ego car's goal in the road topology and the object interactions. This information respects the lanes changed by the ego car and the conditions of object interactions (see Section~\ref{subsec:basic}, \textbf{``Road Topologies"} and \textbf{``Interaction Types"}).

\section{Baseline SimCopilot}

We introduce a baseline SimCopilot for initializing the study on the NLD simulation of the object interactions. This baseline accounts for translating the object motions, generating the complex object interactions, and generalizing the object interactions across various road topologies.

\begin{figure*}[t!]
\centering
\includegraphics[width=\linewidth]{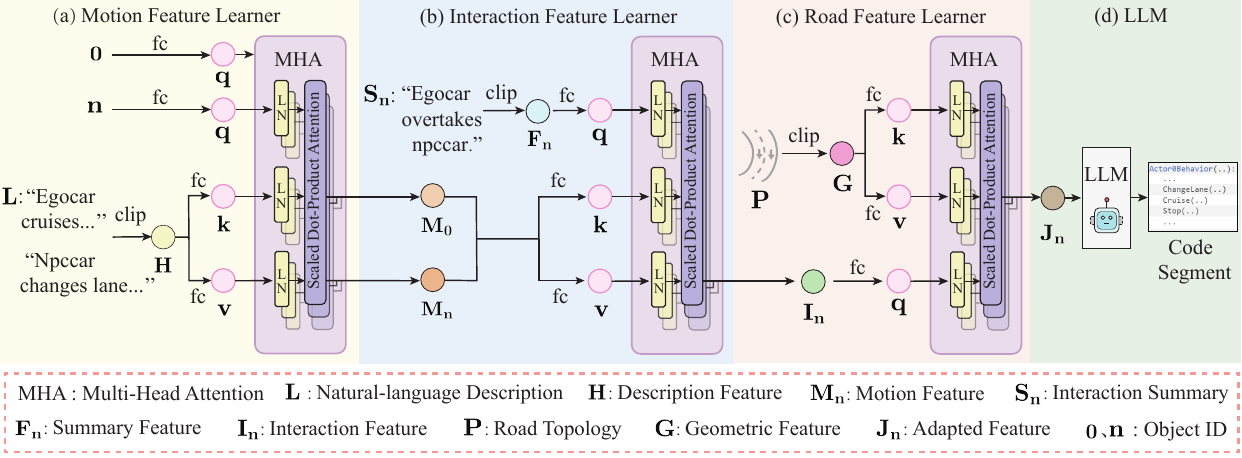}
\vspace{-0.2in}
\caption{The architecture of the baseline SimCopilot. (a) We compute the motion features for the ego car and other obstacles on the same road. (b) With the motion features of the ego car with the index of 0 and the $n^{th}$ obstacle, we compute the interaction feature between them, by the prompt of the interaction summary. (c) We adapt the interaction feature according to the road topology, letting the interaction be reasonable and computing the adapted feature. (d) LLM takes input as the adapted feature to generate the code segment.}
\label{fig:llm}
\vspace{-0.2in}
\end{figure*}

As illustrated in Figure~\ref{fig:llm}, the baseline SimCopilot takes input as the natural-language description of the object motions, the summary of object interactions, and the road topology. It incorporates the motion, interaction, and road feature learners, which assist LLM in generating the code segment for simulating the object interactions.

\vspace{0.05in}
\noindent{\bf Motion Feature Learner~~}
The motion feature learner employs the CLIP text encoder~\cite{radford2021learning}, which embeds the natural-language motion description ${\bf L} \in \mathbb{R}^K$ ($K$ is the number of tokens) of the motions of the ego car and $N$ obstacles, into the description feature ${\bf H} \in \mathbb{R}^C$ ($C$ is the number of feature channels). Next, this learner uses the multi-head attention ($mha$) to extract a set of motion features $\{{\bf M}_n \in \mathbb{R}^C ~|~n=0,...,N \}$ from the latent feature ${\bf H}$. The motion feature ${\bf M}_n$ is computed as:
\begin{equation}
\left\{
\begin{aligned}
    \begin{array}{l}
    \hspace{-1ex}{\bf H} = clip({\bf L}) \vspace{2ex}, \\
    \hspace{-1ex}{\bf q} = fc(n),~{\bf k} = fc({\bf H}),~{\bf v} = fc({\bf H}) \vspace{2ex}, \\
    \hspace{-1ex}{\bf M}_n = mha({\bf q}, {\bf k}, {\bf v}).
\end{array}
\end{aligned}
\right.
\label{eq:motion_learner}
\end{equation}
In Eq.~\eqref{eq:motion_learner}, the fully connected layer $fc$ processes the index of the ego car ($n=0$) or the obstacles ($n \ge 1$) to achieve the query vector ${\bf q} \in \mathbb{R}^C$. We compute the key and value vectors ${\bf k}, {\bf v} \in \mathbb{R}^C$ based on ${\bf H}$. With the query vector ${\bf q}$, the multi-head attention harvests the motion information of the $n^{th}$ object from the description ${\bf L}$. This motion information is represented by ${\bf M}_n$.

\vspace{0.05in}
\noindent{\bf Interaction Feature Learner~~}
The interaction feature learner uses the CLIP text encoder to extract the summary features $\{{\bf F}_n \in \mathbb{R}^C ~|~n=1,...,N \}$ from the interaction summary $\{{\bf S}_n \in \mathbb{R}^K ~|~n=1,...,N \}$. ${\bf S}_n$ with $K$ tokens briefs the interaction between the ego car and the $n^{th}$ obstacle. We use the feature ${\bf F}_n$ to compute the query vector ${\bf q}$, for further computing the interaction feature ${\bf I}_n \in \mathbb{R}^C$ as:
\begin{equation}
\left\{
\begin{aligned}
    \begin{array}{l}
    \hspace{-1ex}{\bf F}_n = clip({\bf S}_n)\vspace{2ex}, \\
    \hspace{-1ex}{\bf q} \!=\!\! fc({\bf F}_n),{\bf k} \!=\!\! fc({\bf M}_0,{\bf M}_n),{\bf v} \!=\!\! fc({\bf M}_0,{\bf M}_n) \vspace{2ex}, \\
    \hspace{-1ex}{\bf I}_n = mha({\bf q}, {\bf k}, {\bf v}).
\end{array}
\end{aligned}
\right.
\label{eq:interaction_learner}
\end{equation}
In Eq.~\eqref{eq:interaction_learner}, we calculate the key and value vectors ${\bf k}, {\bf v}$ based on the motion features ${\bf M}_0,{\bf M}_n$ of the ego car and the $n^{th}$ obstacle. With ${\bf q}$, the multi-head attention focuses on the motions of the ego car and the $n^{th}$ obstacle that leads to the specified interaction. When the motion description is unavailable, we use summary feature ${\bf F}_n$ in place of the interaction feature ${\bf I}_n$.

\vspace{0.05in}
\noindent{\bf Road Feature Learner~~}
The road feature learner also uses the CLIP text encoder to extract the geometric information from the XODR file that globally records the road topology ${\bf P} \in \mathbb{R}^K$, forming the geometric feature ${\bf G} \in \mathbb{R}^C$. In Eq.~\eqref{eq:road_learner}, we use the interaction feature ${\bf I}_n$ to compute the query vector ${\bf q}$, which is fed into the multi-head attention. This learner adapts the global geometric information of the road topology, attending to more local information associated with the interaction between the ego car and the $n^{th}$ obstacle. It yields the adapted feature ${\bf J}_n \in \mathbb{R}^C$ as:
\begin{equation}
\left\{
\begin{aligned}
    \begin{array}{l}
    \hspace{-1ex}{\bf G} = clip({\bf P})\vspace{2ex}, \\
    \hspace{-1ex}{\bf q} = fc({\bf I}_n),~{\bf k} = fc({\bf G}),~{\bf v} = fc({\bf G}) \vspace{2ex}, \\
    \hspace{-1ex}{\bf J}_n = mha({\bf q}, {\bf k}, {\bf v}).
\end{array}
\end{aligned}
\right.
\label{eq:road_learner}
\end{equation}
We input the adapted feature ${\bf J}_n$ to LLM, which is fine-tuned with low-rank adaptation~\cite{hu2021lora}. LLM generates the code segment for simulation.

\section{Challenges and Evaluations}

We train and test the baseline SimCopilot and its ablation variants on the tasks of the translation of object motions, the generation of complex interactions, and the interaction generalization across different road topologies.

\subsection{Translation of Object Motions~~}

\noindent{\bf Task Definition and Evaluation Metric~~}
SimCopilot translates the natural-language description to the motion functions (i.e., \emph{Cruise}, \emph{ChangeLane}, and \emph{Stop}) and parameters (i.e., \emph{direction}, \emph{distance}, \emph{initial velocity}, \emph{accelerated velocity}, and \emph{duration time}) in the code segment. The simulator uses the code segment to control the motions of the ego car and obstacles. Each object has a trajectory in the virtual road scene.

To evaluate the accuracy of translating the object motions, we compute the discrepancy ${\bf T} \in \mathbb{R}$ between the trajectories translated from the natural language and the ground-truth trajectories given in the dataset as:
\begin{align}
{\bf T} =  \sum^{T}_{t=0} \frac{||{\bf E}^t-\widehat{{\bf E}}^t||^2_2}{T} + \sum^{T}_{t=0}\sum^{N}_{n=1}\frac{||{\bf O}_n^t-\widehat{{\bf O}}_n^t||^2_2}{TN}.
\label{eq:similarity}
\end{align}
In Eq.~\eqref{eq:similarity}, we denote ${\bf E}^t, \widehat{{\bf E}}^t \in \mathbb{R}^2$ as the translated and ground-truth 2D positions, respectively, of the ego car's $t^{th}$ waypoint. ${\bf O}_n^t, \widehat{{\bf O}}_n^t \in \mathbb{R}^2$ are the translated and ground-truth 2D positions of the $n^{th}$ obstacle's $t^{th}$ waypoint. The waypoints are uniformly distributed along the trajectory. A smaller discrepancy ${\bf T}$ means the translated and ground-truth trajectories are similar, thus prompting an accurate translation of the object motions.

\vspace{0.05in}
\noindent{\bf Training and Testing Splits~~}
We split the L2I dataset into training and testing sets in translating the object motions. Each set contains 60,000 scenes, where the interactions between the ego car and 1$\sim$5 obstacles appear in 6 kinds of road topologies. We let interaction types be roughly uniform in the training and testing sets.

\vspace{0.05in}
\noindent{\bf Experimental Results~~}
%
In Table~\ref{tab:comparison_motion}, we test the baseline SimCopilot with or without the motion feature learner, reporting the trajectory discrepancy (see Eq.~\eqref{eq:similarity}) in the last column. We build the baseline SimCopilot on top of Llama~\cite{touvron2023llama} with 13B parameters. The motion feature learner improves  SimCopilot by reducing the discrepancy between the generated trajectories and their ground-truth counterparts. SimCopilot evaluated in Table~\ref{tab:comparison_motion} imperfectly translates some motion parameters (e.g., velocity and acceleration), thus leading to many failure cases of controlling the object motions. We comprehensively discuss the failure cases of motion translation in the supplementary file.


\setlength{\tabcolsep}{13pt}
\renewcommand{\arraystretch}{1.2}
\begin{table}
\begin{center}
\begin{tabular}{c|c}
\hline
{\bf Method} &  {\bf Discrepancy}$\downarrow$\\
\hline\hline
{\bf w/o Motion Feature Learner} &   108.8 \\
\hline
{\bf w/ Motion Feature Learner} &  {\bf 103.0}   \\
\hline
\end{tabular}
\end{center}
\vspace{-0.2in}
\caption{The ablation study on the SimCopilot with/without the motion feature learner. We report the performances in terms of trajectory discrepancy.}
\label{tab:comparison_motion}
\end{table}

\subsection{Generation of Complex Interactions~~}

\noindent{\bf Task Definition and Evaluation Metric~~}
In generating complex interactions, we train SimCopilot on the interactions between a few objects. Next, we test SimCopilot by employing it to generate complex interactions between more objects.

The primary setting of the NLD simulation allows the detailed description of object motions and the summary of object interactions to prompt the generation of object interactions. Someone may put much effort into writing a detailed description when the ego car frequently interacts with many surrounding obstacles. The motion parameters must be carefully checked, ensuring that complex interactions happen as expected. In some scenarios where specific details like object positions and velocities are unnecessary, the summary is used alone to generate complex interactions.

\setlength{\tabcolsep}{5pt}
\renewcommand{\arraystretch}{1.2}
\begin{table*}[t!]
\begin{center}
\begin{tabular}{c|c|c|c|c|c|c|c}
\hline
\multicolumn{2}{c}{{\bf 1 v.s. 2$\sim$5}} & \multicolumn{2}{|c}{{\bf 1$\sim$2 v.s. 3$\sim$5}} & \multicolumn{2}{|c}{{\bf 1$\sim$3 v.s. 4$\sim$5}} & \multicolumn{2}{|c}{{\bf 1$\sim$4 v.s. 5}} \\\hline
\multicolumn{1}{c}{{\bf Distance}$\downarrow$} & \multicolumn{1}{|c}{{\bf Rate}$\uparrow$} & \multicolumn{1}{|c}{{\bf Distance}$\downarrow$} & \multicolumn{1}{|c}{{\bf Rate}$\uparrow$} & \multicolumn{1}{|c}{{\bf Distance}$\downarrow$} & \multicolumn{1}{|c}{{\bf Rate}$\uparrow$}& \multicolumn{1}{|c}{{\bf Distance}$\downarrow$} & \multicolumn{1}{|c}{{\bf Rate}$\uparrow$}\\\hline\hline
73.1 $\to$ {\bf 72.0} & 0.92 $\to$ {\bf 1.0} & 49.35 $\to$ {\bf 39.6} & 0.84 $\to$ {\bf 0.86} & 96.27 $\to$ {\bf 81.4} & 0.56 $\to$ {\bf 0.63} & 107.17 $\to$ {\bf 96.6} & 0.35 $\to$ {\bf 0.39} \\
\hline
\end{tabular}
\end{center}
\vspace{-0.22in}
\caption{The ablation study on the interaction feature learner. We use $\to$ for showing the performance change from the SimCopilot without to with the interaction feature learner. We report the performances in terms of the difference between the interaction positions and the success rate of interaction.}
\label{tab:comparison_interaction}
\end{table*}

Given the detailed description of object motions and the summary of object interactions as the input of SimCopilot, the ego car, and the obstacle should interact with each other at the specified position. We compute the distance ${\bf D} \in \mathbb{R}$ between the generated and ground-truth positions of the ego car and the obstacle that interact with each other. We calculate the the distance ${\bf D}$ as:
\begin{align}
&~~~~~~~~~~~~{\bf D} = \sum^{N}_{n=1} \frac{||{\bf E}_n-\widehat{{\bf E}}_n||_2^2+||{\bf O}_n-\widehat{{\bf O}}_n||_2^2}{N}, \nonumber\\
& s.t.~~({\bf E}_n, {\bf O}_n) \in {\bf S}_b \cup {\bf S}_o \cup {\bf S}_y, ~~||{\bf E}_n-{\bf O}_n||^2_2 \le 10,\nonumber\\
&~~~(\widehat{{\bf E}}_n, \widehat{{\bf O}}_n) \in {\bf T}_b \cup {\bf T}_o \cup {\bf T}_y, ~~||\widehat{{\bf E}}_n - \widehat{{\bf O}}_n||^2_2 \le 10.
\label{eq:success_distance}
\end{align}
In Eq.~\eqref{eq:success_distance}, we denote ${\bf T}_b, {\bf T}_o, {\bf T}_y$ as the sets of the pairs of 2D positions. The pair of 2D positions $({\bf E}_n, {\bf O}_n)$ are associated with the ego car and the $n^{th}$ obstacle. They represent the ground-truth positions, where the ego car bypasses/overtakes/yields the obstacle. The sets ${\bf S}_b, {\bf S}_o, {\bf S}_y$ also contain the pairs of 2D positions. $(\widehat{{\bf E}}_n, \widehat{{\bf O}}_n)$ represent the 2D positions where the ego car successfully bypasses/overtakes/yields the $n^{th}$ obstacle. To count the successful interactions, we add separate listeners to the simulator to monitor the interactions between the ego car and every obstacle. A smaller distance means a successful generation.

Given the summary of object interactions only, the ego car and the obstacles should have the specified types of interactions. Whereas the positions of interactions are unnecessary, the ground-truth positions are defined in Eq.~\eqref{eq:success_distance}. In this setting, we calculate the success rate ${\bf R} \in [0,1]$ of the generated interactions as:
\begin{align}
{\bf R} = \frac{|{\bf S}_b \cup {\bf S}_o \cup {\bf S}_y|}{|{\bf T}_b \cup {\bf T}_o \cup {\bf T}_y|},
\label{eq:success_rate}
\end{align}
where $|*|$ is the cardinal of the set. A higher success rate indicates a better performance of SimCopilot.

\vspace{0.05in}
\noindent{\bf Training and Testing Splits~~}
To evaluate the generation of complex interactions, we increase the training data from the interactions between the ego car and fewer obstacles. Then, we use SimCopilot to generate the interactions with more obstacles. Thus, the training v.s. testing splits can be 1 v.s. 2$\sim$5 obstacles (24,000 v.s. 96,000 scenes), 1$\sim$2 v.s. 3$\sim$5 obstacles (48,000 v.s. 72,000 scenes), 1$\sim$3 v.s. 4$\sim$5 obstacles (72,000 v.s. 48,000 scenes), and 1$\sim$4 v.s. 5 obstacles (96,000 v.s. 24,000 scenes).

\vspace{0.05in}
\noindent{\bf Experimental Results~~}
In Table~\ref{tab:comparison_interaction}, we evaluate the baseline SimCopilot's ability to generate complex object interactions. Following the above training and testing splits, we report the distance between the generated and ground-truth positions, where the object interactions happen. Note that SimCopilot generates interaction positions based on the detailed description of the object motions. We also report the success rate of generating the object interactions when SimCopilot is given the interaction summary as input. SimCopilot, with the interaction feature learner in each split, generally improves the performance of generating complex interactions. By comparing the results across different splits, we find that the baseline SimCopilot with the current LLM backbones is still unreliable for generating the complex interactions between more objects, given too simple interactions as the training samples (see the splits of 1 v.s. 2$\sim$5 and 1$\sim$2 v.s. 3$\sim$5). It possibly hints that the interactions between the ego car and more objects contain critical context, which is hard to learn by the baseline SimCopilot from fewer objects' interactions. The supplementary file discusses the baseline SimCopilot's limitation of generating complex interactions.

\subsection{Interaction Generalization across Roads~~}

\noindent{\bf Task Definition and Evaluation Metric~~}
The interaction generalization requires SimCopilot, which is trained on the object interactions in the given road topologies, to generate reasonable interactions in the unseen topologies. We follow the evaluation metrics defined in Eqs.~\eqref{eq:success_distance} and \eqref{eq:success_rate} by using the summary of object interactions and/without the detailed description of object motions.

\vspace{0.05in}
\noindent{\bf Training and Testing Splits~~}
We divide 120,000 scenes evenly into the training and testing sets, where the shapes and lanes of the road topologies are different. Here, we let the number of lanes of each road topology in the testing set double the lanes of the same kind of topology in the training set. It allows SimCopilot to be trained on simple topologies and evaluated on more complicated ones.

\vspace{0.05in}
\noindent{\bf Experimental Results~~}
In Table~\ref{tab:comparison_road}, we experiment with using the baseline SimCopilot, which is trained on the object interactions happening in a fixed set of road topologies, to generate the object interactions in the unseen road topologies. We report the distance of interaction positions of generation and ground truth, along with the success rate of interaction generation, in Table~\ref{tab:comparison_road}, to compare the performance of SimCopilot with or without the road feature learner. The road feature learner achieves the specific information from the input road topology. This information assists the LLM backbone in generating the appropriate interactions by respecting the geometric properties of the given road topology. Yet, the baseline SimCopilot produces a few off-road interactions whose positions are outside the road's boundary. This is because the positional restriction of the generated interactions is unavailable. The supplementary file further discusses the off-road interactions and the possible direction for addressing this problem.


\setlength{\tabcolsep}{10pt}
\renewcommand{\arraystretch}{1.2}
\begin{table}[]
\begin{center}
\begin{tabular}{c|c|c}
\hline
{\bf Method}&{\bf Distance}$\downarrow$ & {\bf Rate}$\uparrow$\\\hline\hline
w/o Road Feature Learner & 167.1 & 0.23  \\\hline
w/ Road Feature Learner & {\bf 155.2} & {\bf 0.25} \\
\hline
\end{tabular}
\end{center}
\vspace{-0.22in}
\caption{The ablation study on the road feature learner. We report the results in terms of the difference between the interaction positions and the success rate of interaction.}
\label{tab:comparison_road}
\end{table}
\section{Conclusion}

The simulation of short-range object interactions is a profound area. It produces data for training and testing autonomous vehicles, which thus have a delicate sense of the surrounding objects and safely interact with them to reduce the risk of traffic crashes. The traditional simulation engines with visual and programming interfaces require users to carefully direct the object interactions according to the individual road scenes at the cost of substantial human labor. In contrast to the visual and programming interfaces, we propose a novel idea of employing the advanced LLMs, enabling the language interface to simulate the short-range object interactions. This simulation is thus natural-language-driven, allowing tremendous data of object interactions on the road scenes with various shapes to be quickly produced. We collect and contribute the L2I benchmark dataset publicly to the AI community to boost the research on the natural-language-driven simulation of short-range object interactions. We also propose a baseline architecture with LLM, SimCopilot, for inspiring prospective ideas in the relevant areas. Apart from controlling the object motions, generating complex interactions, and generalizing interactions across road topologies, which are mainly considered in this paper, we will investigate more practical usage of SimCopilot and the L2I dataset in the future. 
\section{Language-to-Interaction Dataset}


\begin{figure*}[h!]
\centering
\includegraphics[width=\linewidth]{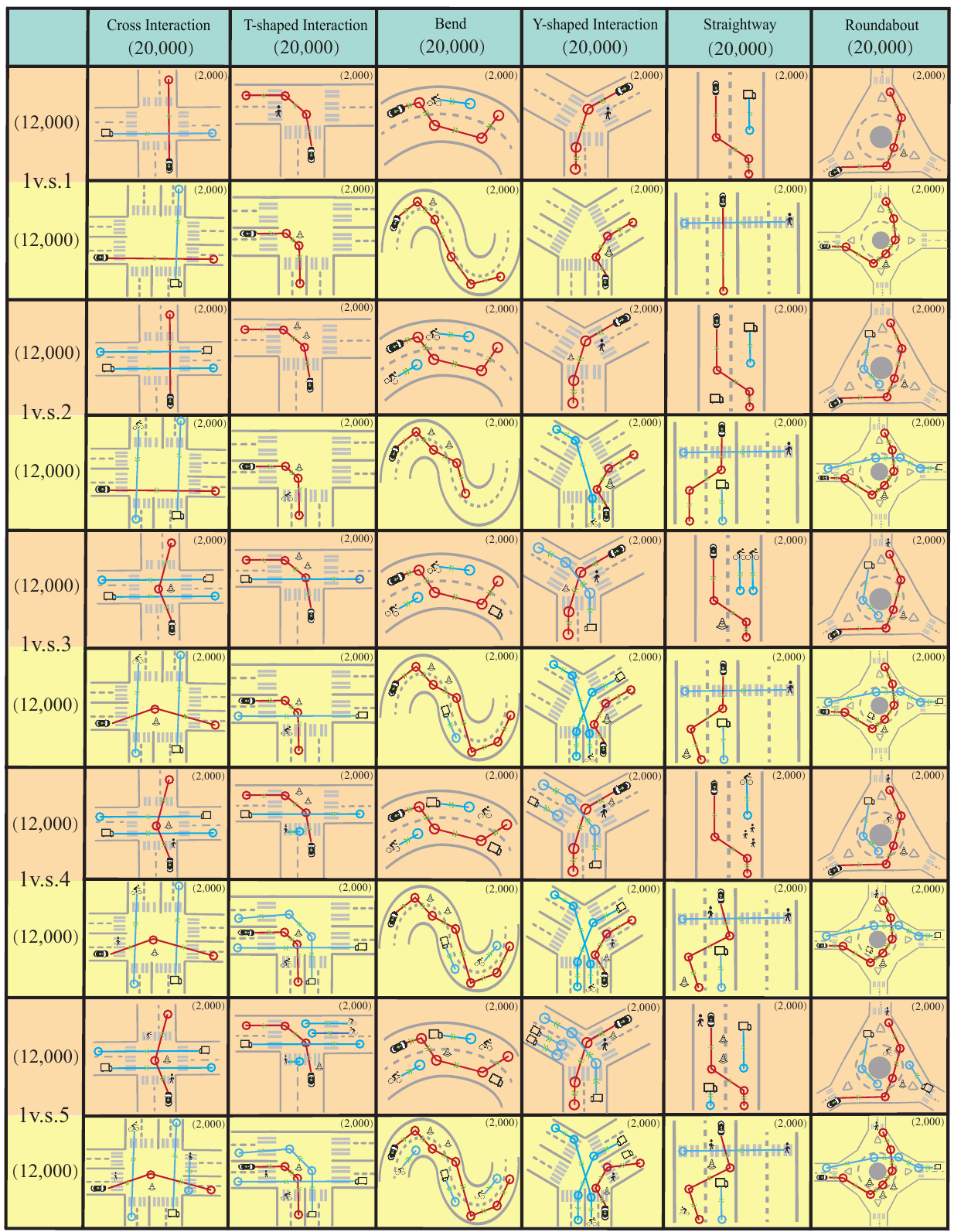}
\vspace{-0.3in}
\caption{More examples of the L2I dataset.}
\label{fig:dataset}
\vspace{-0.15in}
\end{figure*}

In Figure~\ref{fig:dataset}, we provide a statistics of the L2I dataset, which contains 120,000 virtual road scenes. Here, each grid of Figure~\ref{fig:dataset} contains an example of the road scene, the interacting objects and their waypoints, which are show from the bird's eye view.

Along the horizontal direction of Figure~\ref{fig:dataset}, we divide the L2I dataset into six typical kinds of road topology in the L2I dataset, including straightway, bend, roundabout, cross intersection, T-shaped intersection, and Y-shaped intersection. Each kind of road topology has 20,000 examples. It also has two variants with discrepant shapes and lanes, where each variant appears in 10,000 scenes.

Along the vertical direction of Figure~\ref{fig:dataset}, we divide the dataset according to the number of obstacles interacting with the ego car (i.e., 1 v.s. 1, 1 v.s. 2, 1 v.s. 3, 1 v.s. 4, and 1 v.s. 5). Each number of obstacles corresponds to 24,000 scenes, which is evenly divided for training and testing.

\section{Analysis of Failure Cases}

\subsection{Failure Cases of Motion Translation}

We provide the failure cases of motion translation in the first section of the attached video \textbf{``Failure Cases.mp4"\footnote{\url{https://drive.google.com/file/d/1lpiMZxZGv--AYkxuW38f-v-IC4yL1SWZ/view?usp=drive_link.}}} , where the ego car collides with the obstacles. Most of these cases are led by the inaccurate translation between the motion parameters from the natural-language description to the code segment. This may be because the motion parameters occupy a relatively small portion of the natural-language description. During the training of Simcopilot, most of the attention is paid to the translation of the motion names from natural language to code-based functions rather than the motion parameters. A possible way of improving the motion translation can be setting a separate loss (e.g., L2 or L1-smooth loss) for a more focused penalization of the inaccurate translation of motion parameters.

\subsection{Failure Cases of Interaction Generation}

A typical failure case of complex interaction generation can be found in the second section of the attached video \textbf{``Failure Cases.mp4"}. Currently, given too simple interactions as the training samples (see the splits of 1 v.s. 2$\sim$5 and 1$\sim$2 v.s. 3$\sim$5), the baseline SimCopilot with the current LLM backbones is still unreliable for generating the complex interactions between the ego car and more obstacles. Most of the failure cases stem from the missing of the obstacles that are expected to appear as the natural-language description. We conjecture this is because the interactions between the ego car and more objects contain critical context hard to learn by the baseline SimCopilot from fewer objects’ interactions. A solution to address the above problem is to decouple the interaction between the ego car and each obstacle from the description. This solution allows SimCopilot to learn and generate the decoupled interactions, which can be combined to form more complex interactions between more objects.

\subsection{Failure Cases of Interaction Generalization}

We prepare the third section of the attached video \textbf{``Failure Cases.mp4"} to present the failure cases of interaction generalization across road topologies. These failure cases are mainly led by the off-road driving or the cross of solid lines, which belong to the forbidden behaviours in the traffic scenes. Especially, given the interaction summary as a brief prompt, SimCopilot lacks the specify motion parameters, only depending on the motion patterns learned from the seen road topologies, to control the object interactions on the unseen roads. In the future, we plan to involve the negative samples with forbidden behaviours into the training of SimCopilot. By learning from these negative samples, SimCopilot may learn the common knowledge to regularize the object behaviours across different roads.

\section{Application}

The training on the L2I dataset allows SimCopilot to learn the motion patterns of cars on different roads. The trained SimCopilot can generate object interactions in traffic scenes where the ego car bypasses/overtakes/yields static or moving obstacles. The generation further gives us tremendous data to augment other datasets for the downstream perception tasks (e.g., vehicle detection and trajectory prediction). This supplementary file chooses the trajectory prediction as the downstream task to evaluate the usefulness of the data generated by SimCopilot.

Specifically, we employ SimCopilot to generate object interactions in a set of road scenes containing about 300K moving cars. Each object has a trajectory rendered and recorded by the simulator. We use all trajectories to augment the public NGSIM dataset~\cite{ngsim}, which contains the car trajectories captured on the highways, for the trajectory prediction. Here, we use the trajectory data to train the PiP network~\cite{song2020pip}. PiP is pre-trained on about 7M trajectories from NGSIM. We test the performance of PiP in terms of the Root Mean Squared Error (RMSE) metric and the Negative Log-Likelihood (NLL). RMSE measures the predicted trajectory with the maximal maneuver probability, while NLL measures the difference between the predicted trajectory and the ground truth. We fine-tune the pre-trained PiP on the 300K trajectories generated by SimCopilot for two epochs by minimizing RMSE and five epochs by minimizing NLL, with a learning rate of 5e-5.

In Table~\ref{tab:augmentation}, we evaluate the effect of our generated data on the performance of PiP. One should note that the generated trajectories lead to short-range interactions between cars, which are relatively novel compared to the typical behaviors of cars in the NGSIM dataset. These generated trajectories bring diversity to the training data, thus improving the performance of PiP on the trajectory prediction.

\setlength{\tabcolsep}{2.3pt}
\renewcommand{\arraystretch}{1.2}
\begin{table}[]
    \centering
    \begin{tabular}
    {c| c |c c c c c c}
    \hline
         \textbf{Metric}&\textbf{Augmentation}&\textbf{1s}&\textbf{2s}&\textbf{3s}&\textbf{4s}&\textbf{5s}&\textbf{Mean}\\
     \hline
     \hline
     \multirow{2}{*}{\textbf{RMSE}}&\scalebox{0.75}{\XSolidBrush}&0.55&1.23&2.00&2.98&4.27&2.21\\
                     &{\scalebox{0.75}{\CheckmarkBold}}&{\bf 0.54}&{\bf 1.21}&{\bf 1.97}&{\bf 2.93}&{\bf 4.19}&{\bf 2.17}\\
    \hline
    \multirow{2}{*}{\textbf{NLL}}&\scalebox{0.75}{\XSolidBrush}&2.03&3.61&4.44&4.99&5.45&4.10\\
                   \textbf{}&{\scalebox{0.75}{\CheckmarkBold}}&{\bf 1.97}&{\bf 3.56}&{\bf 4.41}&{\bf 4.97}&{\bf 5.44}&{\bf 4.07}\\
     \hline
    \end{tabular}
    \caption{The experiment on the PiP with/without the data augmentation.}
    \label{tab:augmentation}
\end{table}


{\small
\bibliographystyle{ieee_fullname}
\bibliography{egbib}
}





\end{document}